\documentclass[runningheads]{llncs}
\usepackage[T1]{fontenc}
\usepackage{graphicx}
\usepackage{booktabs}
\usepackage[misc]{ifsym}
\newcommand{\corr}{(\Letter)}
\usepackage{microtype}
\usepackage{graphicx}
\usepackage{subfigure}
\usepackage{booktabs} % for professional tables
\usepackage[nolist,nohyperlinks]{acronym}
\usepackage{tikz}
\usetikzlibrary{positioning, shapes, arrows, calc, arrows.meta, patterns}
\usepackage{xspace}
\usepackage{tablefootnote}
\usepackage{array}
\usepackage{makecell}
\usepackage{booktabs}
\usepackage{xspace}
\usepackage{colortbl}
\usepackage{algorithm2e}
\usepackage{amsmath,amsfonts,bm}
\usepackage{todonotes}
\usepackage{multirow}

\usepackage{hyperref}

\makeatletter
\DeclareMathOperator*{\argmin}{arg\,min}
\DeclareRobustCommand\onedot{\futurelet\@let@token\@onedot}
\def\@onedot{\ifx\@let@token.\else.\null\fi\xspace}

\def\ie{\emph{i.e}\onedot}

\makeatother

\usepackage{amsmath}
\usepackage{amssymb}
\usepackage{mathtools}

\newcommand{\ATE}{\textsf{ATE}}
\newcommand{\SMD}{\textsf{SMD}}
\newcommand{\AtwoA}{\textsf{A2A}}

\usepackage{pgfplots}

\pgfplotsset{compat=1.18} 

\usepackage{soul}

\makeatletter
\newcommand{\showspaces}[1]{%
  \begingroup
  \spaceskip=1.5em
  \xspaceskip=1.5em
  \let\@ttfamily\ttfamily
  \def\ {\textvisiblespace}#1%
  \endgroup
}

\begin{document}

\begin{acronym}
    \acro{PSM}{Propensity Score Matching}
    \acro{SMD}{Standardized Mean Difference}
    \acro{ATE}{Average Treatment Effect}
    \acro{NDCG}{Normalized Discounted Cumulative Gain}
    \acro{RF}{Random Forest}
    \acro{LR}{Logistic Regression}
    \acro{CLR}{Chunked Logistic Regression}
    \acro{CV}{cross-validation}
\end{acronym}

\title{Improving Bias Correction Standards by Quantifying its Effects on Treatment Outcomes}

\titlerunning{Improving Bias Correction Standards}
%N.B.: Author information (both in the \author{} and \authorrunning{} command) should only be present in the Camera-Ready Version of your paper. The version that you initially submit for review, ought to be double-blind. So, when initially submitting your paper, use:
%\author{Author information scrubbed for double-blind reviewing}
\author{Alexandre Abraham\inst{1}\orcidID{0000-0003-3693-0560}\corr
\and Andrés Hoyos Idrobo\inst{2}}

%N.B.: comment out the \authorrunning{} command for the double-blind version of your paper submitted for review. Later, if your paper is accepted, use the command for the Camera-Ready Version.

\authorrunning{A. Abraham and A. Hoyos Idrobo}

\institute{Implicity, Paris, France \email{abraham.alexandre@gmail.com} \and
Rakuten Institute of Technology; Rakuten Group, Inc; Paris, France}
\maketitle              % typeset the header of the contribution

\begin{abstract}
With the growing access to administrative health databases, retrospective studies have become crucial evidence for medical treatments. Yet, non-randomized studies frequently face selection biases, requiring mitigation strategies. Propensity score matching (PSM) addresses these biases by selecting comparable populations, allowing for analysis without further methodological constraints. However, PSM has several drawbacks. Different matching methods can produce significantly different Average Treatment Effects (ATE) for the same task, even when meeting all validation criteria. To prevent cherry-picking the best method, public authorities must involve field experts and engage in extensive discussions with researchers.

To address this issue, we introduce a novel metric, A2A, to reduce the number of valid matches. A2A constructs artificial matching tasks that mirror the original ones but with known outcomes, assessing each matching method's performance comprehensively from propensity estimation to ATE estimation. When combined with Standardized Mean Difference, A2A enhances the precision of model selection, resulting in a reduction of up to 50\% in ATE estimation errors across synthetic tasks and up to 90\% in predicted ATE variability across both synthetic and real-world datasets. To our knowledge, A2A is the first metric capable of evaluating outcome correction accuracy using covariates not involved in selection.

Computing A2A requires solving hundreds of PSMs, we therefore automate all manual steps of the PSM pipeline. We integrate PSM methods from Python and R, our automated pipeline, a new metric, and reproducible experiments into \emph{popmatch}, our new Python package, to enhance reproducibility and accessibility to bias correction methods.

\end{abstract}

% Andres: Dude, are you realy using this prompt?
% ChatGPT prompt
% I am a research scientist writing a research paper for the conference called ECML PKDD. I am a french speaking person and I do not speak english perfectly. You are my colleague and you are giving me a hand on this paper. I will give you a text I wrote and I would like you to put it in correct english, by chosing the correct words. I also want you to reformulate in a concise way, as expected from a research paper. Avoid useless adverbs or adjectives, do not use "very" but the right word. Avoid passive formulations and prefer active ones with "we" as a subject. Do not feel obliged to reformulate everything: if part of a sentence is good, keep it. If there is LaTeX code, just leave as it is, do not format it, and if there are acronyms, do not expand them. Just output reformulation, no need to explain anything. Here is the first text to process:

\section{Introduction}

% FINAL

\ac{PSM} seeks to mitigate selection bias between control and treated populations by identifying comparable subpopulations based on potentially confounding covariates. According to the propensity score theorem~\cite{rosenbaum1983central}, under strong ignorability --~i.e., treatment assignment is independent of potential outcomes conditional on the confounding covariates~-- the propensity score can balance treatment groups and allow estimating causal effects.  When applicable, this method is convenient because it  enables matching on a single value rather than full samples, as required in Mahalanobis matching, and it allows for the subsequent use of conventional statistical techniques. Since its initial definition three decades ago~\cite{rosenbaum1983central}, \ac{PSM} has undergone numerous methodological improvements~\cite{stuart2011matchit}, has been proven accurate by meta-analysis~\cite{olmos2014randomized} and therefore recognised as a valid bias correction by French health authorities~\cite{HAS2021}.

Despite these positive signals, \ac{PSM} faces criticism for both theoretical and practical flaws~\cite{king2019propensity}.The typical PSM pipeline~\cite{li2013using}, illustrated in Figure~\ref{fig:pipeline}, includes three steps with multiple options (in blue) and two validation steps (in red). Guidelines recommend progressing as far as possible in the pipeline, making adjustments when validations fail, and backtracking when all alternatives are exhausted~\cite{li2013using}.
Unfortunately, the many options at each step lead to a combinatorial explosion of pipelines, each measuring a different \ac{ATE}, with no established best practices for choosing the best model. Without strict validation, many results may seem valid, potentially misleading practitioners to select pipelines that fit their hypotheses rather than adhering to best practices.

\paragraph{Reporting difficulties.} As each step of this pipeline is manual, the practitioner must carefuly motivates his choices through standardized feature assessment before selection, evaluation of unmatched patients, visual inspection of propensity scores, and more~\cite{heinrich2010primer,li2013using}. This complexity hinders reproducibility and leads to significant shortcomings in method reporting. Systematic reviews of PSM applications reveal that many studies fail to adequately assess the balance of covariate after matching. This issue is observed in 17\% of studies according to a 2008 study~\cite{austin2008critical}, 28\% in 2011~\cite{thoemmes2011systematic}, 26\% in 2015~\cite{mcmurry2015propensity}, 20\% in 2017~\cite{yao2017reporting}, and in a substantial 48\% in a 2020 study~\cite{grose2020use}.
The situation is even more concerning when it comes to reporting the matching method itself with numbers as high as 30\% in a 2008 study~\cite{austin2008critical}, 48\% in 2011~\cite{thoemmes2011systematic}, 33\% in 2015~\cite{mcmurry2015propensity}, and 30\% in a 2017 study~\cite{yao2017reporting}.

Efforts have been made to streamline these best practices, including the creation of visual aids in the MatchIt package~\cite{randolph2014step,lee2017practical}, but ensuring the practitioner's objectivity remains challenging. In this study, we lay the groundwork for a fully automated \ac{PSM} system that, similar to autorank~\cite{herbold2020autorank}, could automatically make decisions and provide a comprehensive report detailing the rationale behind them.

\paragraph{High variability in valid \ac{ATE}.} The first validation step in the pipeline, specific to propensity-based methods, involves \emph{visually inspecting} the propensity score distributions. The second step, common to all bias correction methods, is verifying covariate balance, usually based on \ac{SMD}. Model and hyperparameter choices for propensity estimation and matching can significantly impact variability in the results. This variability can stem from model dependence~\cite{king2019propensity}, unknown confounders, or inaccurate propensity estimation, all of which are difficult to diagnose. For example, MatchIt offers 11 propensity models and 8 matching methods, resulting in 88 possible matchings with varying \ac{ATE}, many of them being valid according to our experiments. Stricter validation may help address this issue, which is the primary goal of this work.

While these issues are still debated , \ac{PSM} remains a reference method in retrospective studies~\cite{yao2017reporting,grose2020use,wang2021use,nowak2022effect}, necessitating a practical response to prevent decisions based on flawed interpretations. We tackle the issue of combinatorial explosion by enhancing method validation to minimize variance in estimated effects. Employing a fully automated pipeline, we investigate multiple PSM methods for each task, highlighting the variability in methods deemed valid by SMD. Then, we demonstrate that our metric A2A aids in reducing their quantity while retaining the most accurate ones. This work is organised as follows. We begin by highlighting the shortcomings of \ac{PSM} assessment. Then, we present our evaluation task and how to create an articifial task from the data at hand and use it to validate a method. Finally, we expose our results and discuss them. Note that the code for all the presented methods and experiments is available in \emph{popmatch}\footnote{Code is available at \url{https://github.com/AlexandreAbraham/popmatch}}, our Python package for PSM.

\tikzstyle{block} = [rectangle, draw, align=center,
    text width=2.6cm, rounded corners, minimum height=4em,
    minimum width=3.0cm]
\tikzstyle{line} = [draw, -{Latex[length=5pt, width=5pt]}]
\tikzstyle{lineback} = [draw, -{Latex[length=5pt, width=5pt]}, dashed]

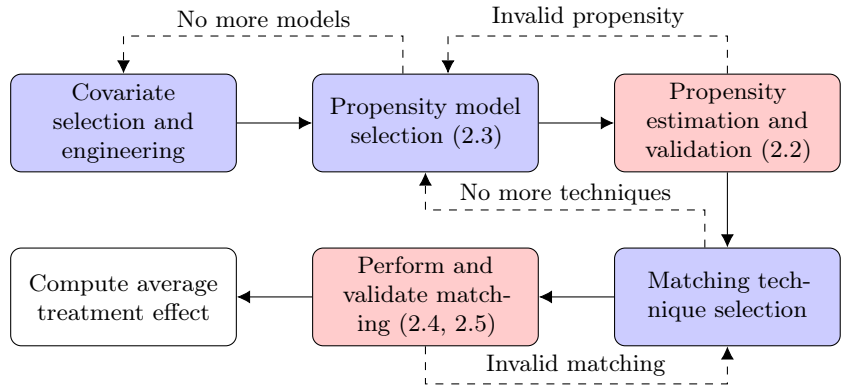
\begin{figure*}
\label{fig:pipeline}
\centering
\begin{tikzpicture}[node distance = 1cm, auto]
    \node [block, fill=blue!20] (init) {Covariate selection and engineering};
    \node [block, fill=blue!20, right=of init] (model) {Propensity model selection (\ref{sec:selection})};
    \node [block, fill=red!20, right=of model] (fit) {Propensity estimation and validation (\ref{sec:propensity})};
    \node [block, fill=blue!20, below=of fit] (technique) {Matching technique selection};
    \node [block, fill=red!20, left=of technique] (matching) {Perform and validate matching (\ref{sec:balancing}, \ref{sec:a2a})};
    \node [block, left=of matching, fill=white] (ate) {Compute average treatment effect};
    \path [line] (init) -- (model);
    \path [line] (model) -- (fit);
    \path [lineback] (fit.north) |- node [midway, above, xshift=-1.85cm] {Invalid propensity} ++(-1.85cm,+.5cm) -| ($(model.north)+(.3cm,0)$);
    \path [lineback] ($(model.north)+(-.3cm,0)$) |- node [midway, above, xshift=-1.85cm] {No more models} ++(-1.85cm,+.5cm) -| (init);
    \path [line] (fit) -- (technique);
    \path [line] (technique) -- (matching);
    \path [line] (matching) -- (ate);
    \path [lineback] (matching.south) |- node [midway, above, xshift=+2.0cm] {Invalid matching} ++(1.8cm,-.5cm) -| (technique);
    \path [lineback] ($(technique.north)+(-.3cm,0)$) |- node [midway, above, xshift=-1.85cm, yshift=-.05cm] {No more techniques} ++(-1.8cm,+.5cm) -| (model);
\end{tikzpicture}
\caption{Propensity score matching pipeline. Blue boxes indicate steps where the practitioner makes choices. Red boxes indicates steps ending with a validation. Backward arrows show points where the practitioner may revisit previous decisions.}
\end{figure*}

\section{Methods}

PSM consists of finding, in a control and treated population, two exchangeable subsets for comparison. For generality, we refer to the two initial sets as  $X_0$ and $X_1$, and their matched counterparts as  $\hat{X}_0 \subseteq X_0$ and $\hat{X}_1 \subseteq X_1$. The associated targets of interest are $Y_0$, $Y_1$, $\hat{Y}_0$, $\hat{Y}_1$. This work focuses on bipartite matching, where all patients from the smaller population are matched with patients from the larger one.
This section outlines our automation of the PSM pipeline and the development of our metric, following the pipeline as presented in Figure~\ref{fig:pipeline}. Instead of experimenting and backtracking like in a single experiment, we 
apply all available options for each task. This approach simulates the variability in results that would occur if different practitioners used different methods for the same study.

\subsection{Real and synthetic tasks}

\paragraph{Synthetic tasks} simulate the treatment effect for each patient, allowing us to make an exact comparison between the measured and real \ac{ATE}. For data generation, we draw inspiration from the setup B in ~\cite{nie2021quasi} and their implementation in the CausalML Python package~\cite{chen2020causalml}. We use synthetic data comprising 3\,000 samples with 10 numerical features and vary the number of confounders from 0 to 10 to observe their impact on the metrics.

\paragraph{Real-life tasks} rely on openly available datasets that include a treatment variable, as summarised in Table~\ref{tab:dataset-description}. Missing data is imputed using the mean for continuous variables and the mode for categorical ones. The objective variable is binary for \emph{Horse Colic} and \emph{NHANES}, and real for \emph{Groupon}. These datasets are used to analyse the behaviour of our metrics and not for performance evaluation since the real ATE is not known.

\begin{table}[bth]
\centering
  \caption{Real-life dataset used for experiments. "Cont." represents continuous features, while "Cat." represents categorical ones.}
  \label{tab:dataset-description}
  \begin{tabular}{>{\raggedright}m{0.13\linewidth}
                  >{\centering\arraybackslash}m{0.15\linewidth}
                  >{\centering\arraybackslash}m{0.10\linewidth}
                  >{\centering\arraybackslash}m{0.10\linewidth}
                  >{\centering\arraybackslash}m{0.25\linewidth}
                  >{\centering\arraybackslash}m{0.15\linewidth}}
    \toprule
    Dataset & \# Samples & \# Cont. & \# Cat. & Treatment & Prediction \\
    \midrule
    Groupon\tablefootnote{Dataset provided by Harry Wang using Groupon data \url{https://www.kaggle.com/code/harrywang/propensity-score-matching-in-python/input}} & 710 & 4 & 2 & Discount applied & Revenue \\
    Horse colic & 300 & 7 & 13 & Surgery or drug & Survival \\
    NHANES\tablefootnote{Provided by Ehsan Karim using openly available data from the CDC \url{https://www.kaggle.com/code/wildscop/propensity-score-matching-on-nhanes}} & 3974 & 4 & 15 & Rheumatoid arthritis & Heart attack \\
    % Add more rows for additional datasets as needed
    \bottomrule
  \end{tabular}
\end{table}

\subsection{Reminder: Computing propensity scores}
\label{sec:propensity}

The propensity score is the probability of receiving treatment given a set of covariates \(\mathbf{x}\):

\begin{equation}
\widehat{\text{PS}}(\mathbf{x}) = \mathbb{P}\left[T = 1 \mid \mathbf{x}\right] + \epsilon,
\end{equation}

where $T$ is a binary treatment indicator, with $T = 1$ if the unit is treated and $T = 0$ otherwise, $\mathbf{x} = (x_1, x_2, \ldots, x_n)$ is a vector of covariates, $\epsilon$ is the error term. Since the predicted probability and its complement can be used as denominators in inverse propensity weighting, it is considered best practice to clip them within the interval $[0.05, 0.95]$~\cite{li2013using}.

Propensity models are trained through feature preprocessing, feature selection, and model optimization. For feature preprocessing, continuous variables are centered and standardized to unit variance, while categorical variables are one-hot-encoded. Feature selection is unnecessary for our synthetic tasks as all features influence either propensity or outcome. Same goes for real tasks as the accompanying documentation suggests no feature is irrelevant.

We consider three propensity models that differ in nature and calibration methods. \textbf{\ac{LR}} is the most common and is calibrated using simple population weighting. \textbf{\ac{CLR}}, introduced in PsmPy~\cite{kline2022psmpy}, is a classic \ac{LR} model using scikit-learn's \ac{LR} with a liblinear backend\footnote{Surprisingly, using a different backend degrades CLR's performance. This behavior warrants further investigation.}. It is calibrated by segmenting the largest population into chunks the size of the smallest one and performing repeated predictions. This distinctive approach, not explained in their paper, yields results different from those of classic \ac{LR}. We also use the more recent \textbf{\ac{RF}}~\cite{zhao2016propensity}, designed to handle non-linear data, and calibrate it using Platt's scaling~\cite{platt1999probabilistic}. For all methods, we consider both raw predicted probabilities and logit link.

\subsection{Automating propensity model hyperparameter selection}
\label{sec:selection}

\begin{figure}[bthp]
\centering
\begin{tikzpicture}[scale=0.7]
  \begin{axis}[
    axis lines=left,
    xmin=0, xmax=1,
    ymin=0, ymax=0.5,
    xlabel={Propensities},
    ylabel={Frequency},
    title={Distributions},
    width=10cm,
    height=6cm,
    legend style={at={(1.0, 1.0)},anchor=north east},
    ybar interval=1.,
    xtick={.05, .15, .25, .35, .45, .55, .65, .75, .85, .95},
    xticklabels={0.05, 0.15, 0.25, 0.35, 0.45, 0.55, 0.65, 0.75, 0.85, 0.95},
    bar width=24pt,
    ]

    \addlegendimage{fill=blue!30}
    \addlegendentry{Distribution 1}
    \addlegendimage{fill=red!30}
    \addlegendentry{Distribution 2}
    \addlegendimage{pattern=north east lines,}
    \addlegendentry{Overlap}

    \addplot+[ybar, fill=blue!30, opacity=.5] plot coordinates {(0.1, 0.05) (0.2, 0.1) (0.3, 0.3) (0.4, 0.22) (0.5, 0.18) (0.6, 0.1) (0.7,0.05)};

    \addplot+[ybar, fill=red!30, opacity=.5] plot coordinates { (0.3, 0.05) (0.4, 0.15) (0.5, 0.28) (0.6, 0.32) (0.7,0.1) (0.8, 0.05) (0.9, 0.05)};

    \addplot+[ybar, opacity=.5, pattern=north east lines, pattern color=violet, ] plot coordinates { (0.3, 0.05) (0.4, 0.15) (0.5, 0.18) (0.6, 0.1) (0.7,0.05)};

  \end{axis}
\end{tikzpicture}
\caption{Example of propensity score histograms for control and treated population. The hatched area corresponds to the overlap between the two.}
\label{fig:overlap}
\end{figure}
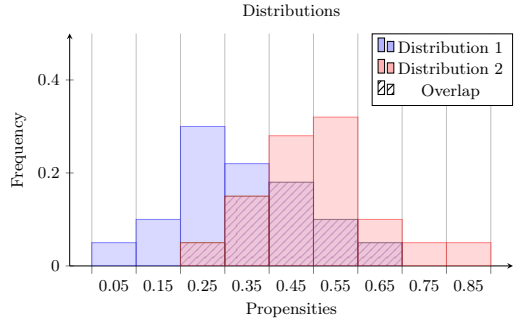

Propensity models may have hyperparameters, such as the number of trees for \ac{RF}. Setting them based on prediction accuracy in a \ac{CV} scheme is impossible since propensity scores lack a ground truth. We propose to use a composite score to evaluate the best method. This score is computed on the left-out test set and is composed of:
\begin{description}
    \item[Model performance] as classification accuracy. Even if the true propensities are unknown, the binary assignment remains relevant information, as we expect the treated group to have a higher propensity:
    $$\text{Accuracy} = \frac{1}{N_0} \big| \{ \mathbf{x} \in X_0 : \widehat{\text{PS}}(\mathbf{x}) < 0.5 \}\big| + \frac{1}{N_1} \big| \{ \mathbf{x} \in X_1 : \widehat{\text{PS}}(\mathbf{x}) \geq 0.5 \}\big|.$$
    
    \item[Extreme value ratio] as the ratio of values outside of the interval $[0.05, 0.95]$ predicted in the left-out test set. While extreme propensities can be clipped, it is preferable to have none:
    $$\text{Extremes} = \frac{1}{N}\Big[ \big| \{ \mathbf{x} \in (X_0 \cup X_1) : \widehat{\text{PS}}(\mathbf{x}) \notin [0.05, 0.95] \}\big|\Big].$$
    
    \item[Overlap coefficient]~\cite{mcgill1979evaluation} between the normalized histograms of the valid propensity scores of both populations in strata of $0.1$. This automates the \textit{visual inspection} recommended in best practices~\cite{li2013using}.
    %\Alex{No, this is not the TV distance.}\Andres{Then, the figure is misleading.}
    $$\textsf{NH}_i(X) = \frac{1}{|X|}\big|\{\mathbf{x} \in X: \widehat{\text{PS}}(\mathbf{x}) \in [i \times 0.1 - 0.05,\, i \times 0.1 + 0.05]\}\big|$$
    $$\text{Overlap} = \sum_{i\in[1, \ldots ,9]} \min(\textsf{NH}_i(X_0),\, \textsf{NH}_i(X_1))$$
    
    This idea is reminiscent of stratified propensity matching and is depicted in Figure~\ref{fig:overlap}. Note that throughout the paper, a propensity method is deemed invalid if this overlap falls below $0.5$.
\end{description}

We use the following composite score in a 5-fold cross-validation (CV) to select the best models:
$$
\text{Score} = \text{Accuracy} + (1 - \text{Extremes ratio}) + \text{Overlap ratio}.
$$

\subsection{Validating matching with covariate balancing}
\label{sec:balancing}

\ac{SMD} computes the distance between two distribution of features, typically control and treated populations. SMD for continuous variables is usually computed using Cohen's D with pooled standard deviation, \ie the T-test effect size. For categorical variables, we use Cramér's V, the $\chi^2$ effect size, as there is no standard:

\begin{equation*}
  \text{Cohen's D} = \frac{{\mu_0 - \mu_1}}{{\sqrt{\frac{{(N_0 - 1) \cdot \sigma_0^2 + (N_1 - 1) \cdot \sigma_1^2}}{{N_0 + N_1 - 2}}}}},  
\quad
    \text{Cramér's V} = \sqrt{\frac{{\chi^2}}{{N \cdot \min(c-1, r-1)}}},
\end{equation*}

with $\mu_0$ (resp. $\mu_1$) the means for the first (resp. second) populations, $N_0$ (resp. $N_1$) their respective cardinalities, $\sigma_0$ (resp. $\sigma_1$) their respective standard deviations, $N$ the total number of observations, and $c$ (resp. $r$) the number of columns (resp. rows) in the contingency table. French Health Authorities consider that having an SMD below 10\% after correction validates the bias correction method. We use the same criterion in our experiments.

\subsection{Contribution: Validating ATE and matching with a new metric}
\label{sec:a2a}

\ac{ATE} is simply the difference between the average effect in both populations.
\begin{equation}
%\ATE(Y_0, Y_1) = \widebar{Y_1} - \widebar{Y_0}
\ATE(Y_0, Y_1) = \mathbb{E}\left[Y_1\right] - \mathbb{E}\left[Y_0\right].
\end{equation}

%\Andres{You didn't include $\widebar{Y_1}$ in the notation. You don't use the bar later. So, why not using the expectation?}

The ability of a \ac{PSM} to retrieve the true ATE can only be assessed using synthetic tasks or through literature on real datasets~\cite{olmos2014randomized}. There exists no method to select the best technique for the specific task being studied.
We propose achieving this by generating artificial matching tasks using the studied data. These tasks replicate biases similar to those in the original task but with known ATE. We posit that an approach's ability to recover the known ATE in these artificial tasks measures its effectiveness in retrieving the ATE in the original task. This procedure evaluates the entire matching process, including the interaction between propensity estimation and matching. We term this new metric \emph{A2A}.

Note that in this manuscript, we make a clear distinction between two types of synthetic problems: \emph{artificial tasks}, which are derived from a homogeneous population and possess a null \ac{ATE}, used for $\AtwoA$ computation, and \emph{synthetic tasks}, where we simulate an entire matching problem with a known ATE to evaluate matching methods. We will maintain this terminology throughout the rest of the manuscript to differentiate between these two types of scenarios.

\subsubsection{Artificial task from real world data}

The core idea of our work is to establish a matching task with a predetermined outcome. We suggest partitioning a sub-population uniformly, for example controls $X_0$, into two subsets $X_0^{(0)}$ and $X_0^{(1)}$. As both subsets originate from the control population, we can ascertain that the treatment effect remains consistent across both, leading to an estimated ATE of zero post PSM.

\paragraph{Selecting subsets.}
To assess the PSM method's efficacy in addressing the reference task, we design our artificial tasks to mimic the biases present in the reference scenario. We select subsets and ensure their unadjusted ATE matches that of the reference task, as shown in Equation~\ref{eqn:ate}, while also enforcing equivalent initial SMD conditions, as illustrated in Equation~\ref{eqn:smd}. 
Since the artificial task uses half the dataset, we introduce only half the biases to avoid making it too challenging and ensure a good spread of scores. This adjustment doesn't impact the final method rankings. Lastly, we determine the cardinality of the two subsets to be proportional to that of the reference task, as indicated in Equation~\ref{eqn:cond}. 

Thus, selecting subsets to match boils down to solving the following optimization problem:

\begin{subequations}
\begin{align}
    \left\{X_0^{(0)}, X_0^{(1)}\right\} \gets & \argmin_{\{X_0^{(0)}, X_0^{(1)}\}} \mathcal{L}\left(X,\, Y,\, \left\{X_0^{(0)}, X_0^{(1)}\right\}\right), \label{eqn:loss} \\
   \text{s.t. }& {
   X_0^{(0)}, X_0^{(1)} \subseteq X_0, \quad X_0^{(0)} \cap X_0^{(1)} = \varnothing, \quad
   \underbrace{\frac{|X_0|}{|X_1|} = \frac{|X_0^{(0)}|}{|X_0^{(1)}|}}_{\text{Same size ratio}}}. \label{eqn:cond}
\end{align}
\end{subequations}
where, 
\begin{subequations}
\label{eqn:main_loss}
\begin{align}
    \mathcal{L}\left(X,\, Y,\, \left\{X_0^{(0)}, X_0^{(1)}\right\}\right)=& \left(\frac{1}{2}\,\underbrace{\ATE(Y_0,\, Y_1)}_{\text{reference task}} - \underbrace{\ATE\left(Y^{(0)}_0,\, Y^{(1)}_0\right)}_{\text{artificial task}}\right)^2 \label{eqn:ate} \\
    &+ \left(\frac{1}{2}\,\underbrace{\SMD(X_0,\, X_1)}_{\substack{\text{reference task}}} - \underbrace{\SMD\left(X^{(0)}_0,\, X^{(1)}_0\right)}_{\substack{\text{artificial task}}}\right)^2. \label{eqn:smd}
\end{align}
\end{subequations}

%\Andres{Also, it has its own set of parameters. Do we relate this to ``stability''?. You mention this before. Do we have a number?}
%\Alex{I do not know what you are referring to. I mention stability once to refer to low variance in the results. I do not know how it would apply here.}
%\Andres{hahhahaha, Yup, but stability in statisitcs can be a strong word. Thus, maybe we need to clarify it and say ``low-variance'' results.}
%\Alex{So what, they own the word now? I can fight a statistician when you want, those bastards always come with wrong a-priori!}
%\Andres{hahahahahaa. Ok. But, just to clarify, variance is a property in expectation, stability is a property with high probability (one can go from stability to variance, the other way around is trickier).}

Where $X^{(0)}_0$, resp. $X^{(1)}_0$, represents our artificial control, resp. treated, population. Without losing generality, we assume $|X_0|\geq |X_1|$.
Ultimately, we obtain a task similar to the original but with a known ATE of 0. 

%\Andres{I'm still digesting this part.}
%\Andres{Also, looks like a variation of Diff-in-Diff with synthetic controls. Do you have any comment about this similitude?}
%\Alex{This metric is meant to evaluate bias correction methods. I am not a specialist of diff in diff but in could also be used to evaluate it. As for the intuition, I see no similarity with diff in diff. What I do here is crafting a proxy problem and hope that methods that solve this problem are also able to solve the original one. Because AFAIK there is not theoretical guarantee that this is the case, this is why I use synthetic tasks to prove that it works.}

%We assigned a weight of $\frac{1}{2}$ to the original SMD and ATE because our artificial problem has a smaller cardinality and potentially different population distribution support. This weight yielded more spread out values for A2A in our experiments, but its specific value does not affect the final ranking of methods. 

This set building task is easier to solve than PSM and there as it allows for the free swapping of samples between populations. Thus it does not require sophisticated methods.
Due to the absence of a well-defined gradient in the optimization problem, 
we use a hill-climbing method for minimization, as detailed in Algorithm~\ref{alg:clustering}. 
In our experiments, the algorithm converged within seconds, making a more complex approach, \ie, simulated annealing, unnecessary. Also, this method ensures non-overlapping subsets by construction and injects diversity into our artificial tasks as it reaches local minimums, while other methods aim for the global one.

\paragraph{A2A: A new metric to assess the best performing PSM pipeline.}
Identifying the best method involves minimizing the difference in ATE measured on the matched populations, $\hat{Y}^{(0)}_0$ and $\hat{Y}^{(0)}_1$:
\begin{equation}
\label{eqn:a2a}
\AtwoA \coloneqq \left|\ATE\left(\hat{Y}^{(0)}_0,\, \hat{Y}^{(1)}_{0}\right)\right|.    
\end{equation}

We call this metric A-to-A, or A2A for short, reminiscently of the A/B test wording, since both populations used in the task come from the A population. The closer it is to zero, the better. To obtain the A2A score, we compute 100 bootstraps of the entire A2A process, from task creation to ATE estimation, on the larger population, whether control or treated, and take the mean of the resulting values. Unlike SMD, $\AtwoA$ is a relative metric: Lower values indicate better performance, but there is no guarantee that the debiasing problem is solvable, making it impossible to set a fixed threshold like SMD.

%\Andres{
%
%We can effortlessly extend our approach to discrete treatments of higher cardinality, as it implies a larger number of clusters. However, the extension to continuous variables is out of the scope of the paper.
%}
%\Alex{I would not say "effortlessly" as you may face nasty combinatorial explosion. Also I do not understand the reference to "extension to continuous variables".}
%\Andres{You are right. I just wanted to emphasise that it is not that complicated to find a local minima. Yet, the constraints require to check every pairs of subsets.Don't mind the ``continuous'' comment, it is not pertinent.}

\subsubsection{Combining SMD and A2A}
\label{sec:combining}
Since SMD is an absolute value, it allows for the use of a fixed threshold for method selection. In contrast, $\AtwoA$ is inherently relative and cannot use a fixed threshold. In order to combine both metrics in a single method selection process, we tested three strategies. The most straightforward, Min A2A, selects the best performing matching in terms of A2A among those deemed valid by SMD. For fairness, we added its counterpart Min SMD. 
The other two strategies are more permissive, selecting a set of valid PSMs similar to SMD thresholding. SMD\texttimes{}A2A relies on an algorithmic approach, applying DBScan~\cite{ester1996density} to cluster methods based on their SMD and A2A scores. Selecting the best methods boils down to keeping the cluster that contains the PSM with the lowest A2A score. The Pareto method, stemming from multi-objective optimization, selects the Pareto optimal PSM according to both A2A and SMD, among PSM deemed valid by SMD.

% Since SMD is a strict filter and A2A is made to select top performers, our first strategy, called , involves removing invalid matchings according to SMD and then using A2A to select the top remaining ones. 
% The second strategy intertwines both approaches by running A2A's DBScan~\cite{ester1996density} on both metrics, instead of A2A only, to enhance the relevance of the cluster. We then select the cluster containing the solution that is valid with respect to SMD and has the lowest A2A. We call this approach SMD\texttimes{}A2A.

\subsection{Contribution: Popmatch python package}

% FINAL

The most commonly used implementation of PSM algorithms is the R package MatchIt~\cite{stuart2011matchit}. While a Python counterpart exists, PsmPy~\cite{kline2022psmpy}, it includes only a single matching method. To improve reproducibility and broaden accessibility for Python-based data scientists, we introduce \textit{popmatch}, a package that includes the original implementation of PsmPy and a Python interface to the methods offered by MatchIt. The code for downloading tasks, performing PSM selection, method evaluation, and all the experiments in this paper is available and reproducible.

\subsection{Experimental setup}

For each task, as required for A2A computation, we create 100 bootstrapped artificial tasks. We also use these tasks to compute metrics and gain insights, as described in this section. Since our datasets are tabular and our algorithms are easily tractable, all experiments can be completed within a few hours.
%\Andres{What can we say about the machine requirements and computation time.}
%\Alex{Nothing interesting. It's fast and not very demanding.}
%\Andres{Ok, let's make a line on this.}
% Added!

%\subsubsection{Matching methods}
%
Because MatchIt's propensity estimation and matching are tightly linked, we did not separate them. We retained methods that ran without errors on all bootstraps and are closest to 1-to-1 matching, specifically \textit{nearest} and \textit{optimal}, along with propensity score estimation methods \textit{GLM}, \textit{GAM}, \textit{Elasticnet}, \textit{Rpart}, \textit{CBPS}, and \textit{Bart}\footnote{The MatchIt package refers to ``distance'' as a method to estimate propensity scores.}. On the Python side, we combined LR, CLR, and RF propensity estimation methods with PsmPy's matching.

%\Andres{This text is not clear. Is glm a distance?}
%\Alex{"distance" is how methods to estimate propensity scores are called in the Matchit Package.}
%\Andres{Then, let's convert this comment into a footnote.}

\begin{table*}[btp]
  \centering
  \caption{
    Overlap ratio of propensity scores depending on models and datasets. Colored cells indicate scores below $0.5$. Bold numbers indicate the largest overlap row-wise.
    %\Andres{Does it also implies high statistical significance?}\Alex{No, it simply implies a high overlap in propensity scores. Since it's not used we could remove it. Maybe you can do better than me but I found no paper defining what distributions "overlapping enough" were. Usually people show curves and say "yep it overlaps enough". I have chosen the threshold of .5 a bit arbitrarily.}
    %\Andres{It is a fair answer. We just have to mention that.}
  }
  \label{tab:overlap}

\newcolumntype{E}{>{\centering\arraybackslash}m{1cm}}
\begin{tabular}{llrrr@{\hspace{4mm}}rrr}
%{l|EEE|EEE}
\toprule
\multicolumn{2}{c}{Transform} & \multicolumn{3}{c}{None} & \multicolumn{3}{c}{Logit} \\
%\cmidrule(lr){2-4} \cmidrule(lr){5-7}
\midrule
\multicolumn{2}{c}{Model} &       {LR} &   {RF} & {CLR} &    {LR} &   {RF} & {CLR} \\
\midrule
\multicolumn{2}{c}{Groupon}        &    \textbf{ 0.75} & 0.70 &  0.67 & \cellcolor{red!25} 0.25 & \cellcolor{red!25} 0.27 & \cellcolor{red!25} 0.26 \\
\multicolumn{2}{c}{Horse}          &     0.51 & \cellcolor{red!25} 0.29 &  \textbf{0.84} & \cellcolor{red!25} 0.18 & \cellcolor{red!25} 0.28 & \cellcolor{red!25} 0.19 \\
\multicolumn{2}{c}{NHANES}         &    \cellcolor{red!25} 0.47 & \cellcolor{red!25} 0.46 &  \textbf{0.83} & \cellcolor{red!25} 0.12 &\cellcolor{red!25} 0.00 & \cellcolor{red!25} 0.19 \\
\midrule
\multicolumn{2}{c}{Synth. data} &&&&&&\\
\parbox[t]{2mm}{\multirow{11}{*}{\rotatebox[origin=c]{90}{\textit{\# confounders}}}} 
&\, 0  &     \textbf{1.00} & 0.86 &  0.97 & \cellcolor{red!25} 0.00 & 0.62 & \cellcolor{red!25} 0.45 \\
&\, 1  &     \textbf{1.00} & 0.81 &  0.97 & \cellcolor{red!25} 0.00 & 0.61 & \cellcolor{red!25} 0.43 \\
&\, 2  &     0.85 & 0.60 &  \textbf{0.97} & \cellcolor{red!25} 0.36 & \cellcolor{red!25} 0.41 & \cellcolor{red!25} 0.44 \\
&\, 3  &     \textbf{0.99} & 0.73 &  0.96 & \cellcolor{red!25} 0.20 & 0.53 & \cellcolor{red!25} 0.43 \\
&\, 4  &     \textbf{1.00} & 0.87 &  0.97 & \cellcolor{red!25} 0.17 & 0.52 & \cellcolor{red!25} 0.43 \\
&\, 5  &     0.88 & 0.91 &  \textbf{0.95} & \cellcolor{red!25} 0.38 & 0.57 & \cellcolor{red!25} 0.44 \\
&\, 6  &     \textbf{1.00} & 0.86 &  0.97 & \cellcolor{red!25} 0.00 & \cellcolor{red!25} 0.27 & \cellcolor{red!25} 0.40 \\
&\, 7  &     \textbf{1.00} & 0.89 &  0.97 & \cellcolor{red!25} 0.00 & 0.59 & \cellcolor{red!25} 0.44 \\
&\, 8  &     0.88 & 0.83 &  \textbf{0.97} & \cellcolor{red!25} 0.35 & \cellcolor{red!25} 0.40 & \cellcolor{red!25} 0.43 \\
&\, 9  &     0.91 & 0.96 &  \textbf{0.98} & \cellcolor{red!25} 0.38 & 0.76 & \cellcolor{red!25} 0.43 \\
&\, 10 &     0.91 & 0.90 &  \textbf{0.96} & \cellcolor{red!25} 0.39 & \cellcolor{red!25} 0.04 & \cellcolor{red!25} 0.44 \\
\bottomrule
\end{tabular}
\end{table*}

\begin{table}[ht]
  \centering
  \caption{Ranking based correlation between SMD and the amount of correction applied, the ground truth, and random values as sanity check.}
  \label{tab:correlation}
\begin{tabular}{ll rr@{\hspace{4mm}}rr@{\hspace{4mm}}rr}
\toprule
{} & {} & \multicolumn{2}{c}{Correction magnitude} & \multicolumn{2}{c}{Ground truth} & \multicolumn{2}{c}{Random} \\
%\cmidrule(lr){2-3} \cmidrule(lr){4-5} \cmidrule(lr){6-7}
\cmidrule(lr){3-4} \cmidrule(lr){5-6} \cmidrule(lr){7-8}
\multicolumn{2}{c}{Dataset} & \multicolumn{1}{c}{Kendall's $\tau$} & pValue & \multicolumn{1}{c}{Kendall's $\tau$} & pValue & \multicolumn{1}{c}{Kendall's $\tau$} & pValue \\
\midrule
\multicolumn{2}{c}{Groupon}        &  -0.30 ± 0.58 &        0.02 &     0.20 ± 0.45 &        0.05 &  -0.00 ± 0.16 &        0.51 \\
\multicolumn{2}{c}{Horse}          &   0.05 ± 0.30 &        0.34 &     0.05 ± 0.26 &        0.39 &   0.00 ± 0.15 &        0.56 \\
\multicolumn{2}{c}{NHANES}         &  -0.01 ± 0.56 &        0.11 &     0.39 ± 0.28 &        0.16 &   0.01 ± 0.17 &        0.53 \\
\midrule
\multicolumn{2}{l}{Synth. data} &&&&&& \\
\parbox[t]{2mm}{\multirow{11}{*}{\rotatebox[origin=c]{90}{\textit{\# confounders}}}} 
&\, 0  &  -0.39 ± 0.27 &        0.16 &     0.17 ± 0.31 &        0.31 &   0.01 ± 0.18 &        0.48 \\
&\, 1  &  -0.35 ± 0.27 &        0.14 &     0.16 ± 0.26 &        0.33 &   0.01 ± 0.17 &        0.52 \\
&\, 2  &  -0.39 ± 0.22 &        0.11 &     0.18 ± 0.23 &        0.32 &   0.01 ± 0.17 &        0.50 \\
&\, 3  &  -0.40 ± 0.21 &        0.11 &     0.20 ± 0.26 &        0.34 &  -0.02 ± 0.17 &        0.49 \\
&\, 4  &  -0.31 ± 0.31 &        0.14 &     0.15 ± 0.23 &        0.39 &  -0.00 ± 0.17 &        0.48 \\
&\, 5  &  -0.35 ± 0.26 &        0.14 &     0.14 ± 0.20 &        0.41 &   0.01 ± 0.19 &        0.44 \\
&\, 6  &  -0.40 ± 0.26 &        0.11 &     0.19 ± 0.23 &        0.31 &   0.02 ± 0.18 &        0.49 \\
&\, 7  &  -0.30 ± 0.31 &        0.22 &     0.12 ± 0.24 &        0.36 &  -0.02 ± 0.16 &        0.47 \\
&\, 8  &  -0.39 ± 0.31 &        0.15 &     0.11 ± 0.19 &        0.41 &  -0.01 ± 0.17 &        0.50 \\
&\, 9  &  -0.32 ± 0.27 &        0.14 &     0.12 ± 0.23 &        0.33 &   0.02 ± 0.18 &        0.45 \\
&\, 10 &  -0.30 ± 0.30 &        0.18 &     0.08 ± 0.24 &        0.38 &  -0.01 ± 0.17 &        0.49 \\
\bottomrule
\end{tabular}
\end{table}

\section{Results}

\subsection{Validating propensity scores}

As explained in Section~\ref{sec:selection}, propensity estimation is considered valid if the overlap between the two distributions is at least 50\%. Table~\ref{tab:overlap} shows these ratios and indicates whether each method meets the threshold.

Regarding model validity, we first observe that applying the logit link, a recommended measure enabled by default in PsmPy, tends to yield invalid propensity models. Second, PsmPy's CLR achieves remarkable results compared to LR and is the only method able to yield valid models on the Horse and NHANES datasets. Although this is clearly the optimal choice, we keep all models without logit link for the rest of our analysis.

\subsection{Assessing SMD as a performance evaluator}
\label{sec:smd}

Having artificial tasks with known ATE provides an opportunity to evaluate SMD's ability to predict the performance of matching methods. We assess this based on the correlation between SMD's ranking and the ground truth. Additionally, we assess the correlation with the magnitude of the correction applied, \ie, the difference between the uncorrected ATE and the corrected one. We use Kendall's $\tau$ correlation score due to our limited assumptions about the predicted ATE. As a sanity check, we compare it to a random ranking.

Table~\ref{tab:correlation} shows that, as expected, the SMD ranking does not correlate with a random one. 
%The correlation of SMD with the ground truth spans from 0.05 to 0.39, which can be qualified as a weak correlation. However, SMD exhibits a stronger anti-correlation with magnitude since it is around -0.3 for almost all tasks. Given that lower SMD is better, it means that what SMD measures is the amount of correction applied to the data, which is logical. The problem arises when there is an overcorrection as the correction still increases but starts deviating from real ATE.
%
%\Alex{I have completely rewritten this and added results}
Most correlations between SMD and the ground truth are between 0.1 and 0.2, indicating a weak correlation. However, SMD demonstrates a stronger negative correlation with magnitude, revolving around -0.3 across most tasks. As lower SMD values are better, this implies that SMD measures the extent of correction rather than the accuracy or unbiasedness of the final result. This behavior aligns with the expectation that feature differences decrease with correction. However, the challenge is finding the optimal balance, as overcorrection can occur, leading to inaccuracies in the estimated effect.

\subsection{Validity of matching methods}
\label{sec:matching}

Table~\ref{tab:real} presents results from real datasets, illustrating the trade-offs between both metrics. Methods deemed invalid by SMD are marked in red. Our analysis underscores the need for our approach: no single method consistently outperforms all others, making model selection essential.

We also observe that the metrics are complementary. For instance, ElasticNet and Rpart are identified as the best by A2A for Groupon but are invalid by SMD, while GAM and GLM show a similar pattern for Horse. Conversely, GAM and GLM are rated best by SMD for Groupon and NHANES but are not top-rated according to A2A. This trade-off is illustrated in Figure~\ref{fig:smda2a}.

Regarding matching methods, Bipartify stands out as the only method providing valid matches according to SMD for all datasets, although only one model is selected as the best for NHANES. Overall, PsmPy's CLR offers the best trade-off between metrics, despite narrowly failing to have a valid model selected for Horse.

\begin{table}
  \centering
  \caption{
    SMD and A2A for valid matching methods on real datasets.Red cells indicate results considered invalid according to SMD.
    }
  \label{tab:real}
\begin{tabular}{lrrr@{\hspace{4mm}}rrr}
\toprule
{} & \multicolumn{3}{c}{\textsf{SMD}} & \multicolumn{3}{c}{\textsf{A2A}} \\
\cmidrule(lr){2-4} \cmidrule(lr){5-7}
Matching & Groupon & Horse & NHANES &  Groupon & Horse & NHANES \\
\midrule
{\textit{Optimal}} &&&&&& \\
\quad ElasticNet  &  \cellcolor{red!25} 0.118 & \cellcolor{red!25} 0.128 &  0.046 & \cellcolor{red!25}  84.327 & \cellcolor{red!25} 0.036 &  0.002 \\
\quad Rpart      & \cellcolor{red!25}  0.164 & \cellcolor{red!25} 0.151 &  0.070 & \cellcolor{red!25}  88.880 & \cellcolor{red!25} 0.038 &  0.003 \\
\quad CBPS        &   0.052 & \cellcolor{red!25} 0.117 &  0.016 &  118.526 & \cellcolor{red!25} 0.036 &  0.006 \\
\quad Bart        &   0.034 & \cellcolor{red!25} 0.106 &  0.024 &  145.374 & \cellcolor{red!25} 0.053 &  0.003 \\
\quad GAM         &   0.026 & \cellcolor{red!25}0.115 &  0.015 &  220.050 & \cellcolor{red!25} 0.035 &  0.003 \\
\quad GLM         &   0.026 &\cellcolor{red!25} 0.115 &  0.015 &  220.050 & \cellcolor{red!25} 0.035 &  0.003 \\
\midrule
\textit{Bipartify} &&&&&& \\
\quad RF       &   0.051 & 0.077 &  0.044 &  588.262 & 0.043 &  0.002 \\
\quad LR       &   0.039 & 0.086 &  0.032 &  624.086 & 0.048 &  0.001 \\
\quad CLR    &   0.054 & 0.076 &  0.056 &  834.433 & 0.047 &  0.002 \\
\midrule
\textit{PsmPy} &&&&&& \\
\quad CLR        &   0.026 & \cellcolor{red!25} 0.107 &  0.028 &  292.465 & \cellcolor{red!25}0.037 &  0.004 \\
\quad RF           &   0.051 & \cellcolor{red!25}0.179 &  \cellcolor{red!25} 0.166 & 1491.992 & \cellcolor{red!25}0.039 &  \cellcolor{red!25}0.001 \\
\quad LR           &   0.043 & \cellcolor{red!25}0.158 &  \cellcolor{red!25} 0.165 & 1597.057 & \cellcolor{red!25}0.041 & \cellcolor{red!25} 0.003 \\
\bottomrule
\end{tabular}
\end{table}

\subsection{Combining SMD and A2A: A better matching method evaluation}
% Final
\label{sec:smda2a}

As demonstrated, model dependence can result in significant variance in the estimations of models deemed valid. We report this variance as the range between the highest and lowest ATE estimations among valid PSM methods, with a smaller range being preferable. Since Min A2A and Min SMD select only one model each, their range is null. Secondly, we measure the average error of validated models on synthetic tasks with known ground truth. 

Table~\ref{tab:results} demonstrates that SMD\texttimes{}A2A and Min A2A, which prioritize A2A, lead in terms of range and error when the number of confounders is low. As the number of confounders increases, Pareto becomes the best among methods selecting multiple PSMs, while Min SMD consistently outperforms the others.

A significant challenge in bias correction is its one-shot nature, where there is no room for error. Thus, a method's reliability is judged by its weakest performance. Both Min A2A and Min SMD excel in their respective domains, but they falter significantly on Synthetic 3 (Min SMD) and Synthetic 6 (Min A2A). Overall, Pareto strikes the best balance, making it the most reliable PSM selection method.

On real datasets, Pareto achieves the best range for Groupon and slightly improves over SMD for NHANES, while SMD\texttimes{}A2A excels for Horse and NHANES. Based on these observations, we hypothesize that Groupon likely has more confounders.

% that A2A alone is not compelling in both tasks. It has almost all the time a larger range than SMD and highest ATE error, especially when the number of confounders is high. However, combining the two metrics delivers remarkable performance, outperforming SMD in almost all tasks. SMD\texttimes{}A2A performs particularly well on all real tasks yielding the best ranges and if never far from the optimal in terms of ATE error. SMD remains competitive on problem with high number of confounders.

% SMD$\times$A2A is more conservative and only brings a small uplift over SMD on Groupon and Harse, but it leads the way on most synthetic tasks. 

\begin{table}
  \centering
  \caption{Metric-based performance of PSM on real and artificial tasks. \textit{Range of ATE values} represents the difference between the maximum and minimum ATE across valid methods. For synthetic tasks with known ATE, the mean squared error is reported per method. Since Min A2A and SMD select only one PSM, their range is 0 and not shown.
  }
  \label{tab:results}
\begin{tabular}{ll@{\hspace{4mm}}rrr@{\hspace{4mm}}rrr@{\hspace{4mm}}rr}
%{l@{\hspace{2mm}}|@{\hspace{2mm}}r@{\hspace{2mm}}r@{\hspace{2mm}}r@{\hspace{2mm}}r@{\hspace{2mm}}|@{\hspace{2mm}}r@{\hspace{2mm}}r@{\hspace{2mm}}r@{\hspace{2mm}}r}
\toprule
& & \multicolumn{3}{c}{Range of \textsf{ATE} values} & \multicolumn{5}{c}{\textsf{ATE} estimation error} \\
 %\cmidrule(lr){2-4}  \cmidrule(lr){5-9}
 \cmidrule(lr){3-5}  \cmidrule(lr){6-10}
\multicolumn{2}{l}{Dataset} &  \textsf{SMD} & \small\makecell{\textsf{SMD}\\\texttimes\textsf{A2A}} & \makecell{Par-\\eto} & \textsf{SMD} & \small\makecell{\textsf{SMD}\\\texttimes\textsf{A2A}} & \makecell{Par-\\eto} & \makecell{Min\\A2A} & \makecell{Min\\SMD}\\
\midrule
\multicolumn{2}{l}{Groupon}        &       1417 &            1176 &        \textbf{659} &    \multicolumn{5}{c}{\cellcolor{gray!25}} \\
\multicolumn{2}{l}{Horse}          &    0.014 &               \textbf{0.000} &             0.014 &    \multicolumn{5}{c}{\cellcolor{gray!25} \textit{Real ATE unknown}} \\
\multicolumn{2}{l}{NHANES}         &    0.022 &               \textbf{0.000} &             0.005 &    \multicolumn{5}{c}{\cellcolor{gray!25}} \\
\midrule
\multicolumn{2}{l}{Synth. data} &&&&&& \\
\parbox[t]{2mm}{\multirow{11}{*}{\rotatebox[origin=c]{90}{\textit{\# confounders}}}}
& \, 0  &          0.067 &               0.032 &    \textbf{0.031} &         0.054 &              \textbf{0.022} &            0.035 &             0.063 &    \textbf{0.006} \\
&\, 1  &          0.099 &      \textbf{0.003} &    \textbf{0.003} &         0.070 &              \textbf{0.043} &            0.045 &     \textbf{0.039} &            0.052 \\
&\, 2  &          0.099 &      \textbf{0.013} &             0.013 &         0.063 &              \textbf{0.043} &            0.046 &     \textbf{0.030} &            0.058 \\
&\, 3  &          0.122 &      \textbf{0.079} &             0.094 &         0.069 &              \textbf{0.062} &            0.066 &     \textbf{0.028} &            0.103 \\
&\, 4  &          0.096 &               0.046 &    \textbf{0.012} &         0.059 &              0.060 &            \textbf{0.046} &             0.041 &    \textbf{0.039} \\
&\, 5  &          0.087 &               0.064 &    \textbf{0.018} &         0.057 &              0.057 &            \textbf{0.026} &     \textbf{0.013} &            0.039 \\
&\, 6  &          0.095 &      \textbf{0.000} &             0.028 &         0.066 &              0.094 &            \textbf{0.063} &             0.094 &    \textbf{0.049} \\
&\, 7  &          0.067 &               0.043 &     \textbf{0.033} &         0.041 &              0.046 &            \textbf{0.037} &             0.074 &    \textbf{0.030} \\
&\, 8  &          0.046 &               0.029 &    \textbf{0.007} &         0.034 &              0.034 &            \textbf{0.030} &             0.035 &     \textbf{0.025} \\
&\, 9  &          0.079 &               0.070 &    \textbf{0.038} &         0.049 &              0.053 &           \textbf{0.043} &             0.093 &    \textbf{0.019} \\
&\, 10 &          0.078 &               0.038 &    \textbf{0.032} &         0.052 &              0.051 &            \textbf{0.030} &     \textbf{0.023} &            0.044 \\
\bottomrule
\end{tabular}
\end{table}

\section{Discussion}

Applying PSM is not inherently challenging for a trained practitioner. However, the current SMD-based validation is too lenient, allowing multiple methods with varying ATEs to be considered valid. This situation forces practitioners to narrow their method search and justify decisions to authorities, who must understand and evaluate the study, potentially challenging choices due to the absence of established best practices for selecting propensity computation methods. Furthermore, without rigorous reporting standards, errors or data manipulation can go unnoticed, potentially leading to poor decisions.

\paragraph{A2A effectively reduces the variance in estimated ATEs without compromising accuracy.} We have observed that SMD excels at estimating the amount of bias correction applied and measuring ATE, especially in the presence of confounders, where propensity and outcome are explained by the same covariates. However, its weaknesses emerge when the factors explaining the outcome and the selection are disjoint, particularly when the number of confounders is low. In these cases, A2A proves its usefulness through its measurement of ATE on artificial tasks. This difference makes sense if we consider that propensity-based correction are only able to accurately balance covaraites that are linked to selection. In the presence of confounders, correcting propensity often results in well-corrected ATE. However, when ATE is influenced by covariates unrelated to propensity, A2A becomes crucial in selecting models that provide the most accurate ATE correction. This explains A2A's relative nature. While each method excels in its domain, combining them in a Pareto optimal approach ensures a stable model selection method, effective regardless of the number of confounders.

\paragraph{The best PSM pipeline.} While this is not the primary objective of our study, we conducted comparisons of matching methods on our tasks. The main finding is that applying logit on predicted probabilities, or using the emph{nearest} matching in MatchIt, results in models that are considered invalid based on the SMD. PsmPy~\cite{kline2022psmpy} demonstrated superior performance compared to MatchIt~\cite{stuart2011matchit} without identifying a fundamental cause. Our study, which integrates all methods into a single package, reveals that this may be due to CLR, PsmPy's block-based model calibration. CLR shows significantly better overlap in propensities (Table~\ref{tab:overlap}) and improved metric performance on real datasets, particularly Groupon (Table~\ref{tab:real}). This motivates the integration of CLR into MatchIt for further experiments.

\paragraph{Towards an Automated Pipeline and Benchmark for PSM.} This work introduces an automated pipeline capable of performing PSM that satisfies all the requirements of the French Health Authorities. Such a standardized automated baseline could provide a valuable foundation for expert discussions between authorities and applicants. Running it serves as a proof of good faith, and health authorities may request it when further validation is required. This also paves the way for the development of PSM comprehensive benchmarks.

% \textbf{Artificial task for evaluation on studied data.} Synthetic tasks are great but when tackling a real problem, we would like to gain insight on this problem and not a simulation of it. For this purpose we proposed to create artificial tasks where an homogeneous population from the original data is split into two populations biased \wrt each other but where the real ATE is zero. Consequently, the absolute ATE measured on this artificial task is an indication of the PSM performance on this dataset. We call this metric A2A. Table~\ref{tab:results} shows the most striking proof of A2A's validity as, once combined with SMD in SMD+A2A, it greatly narrows the range of ATE prediction, up to 90\%, on our tasks. We also see in Table~\ref{tab:real} that order of difficulty for valid SMD estimation goes from easy NHANES, medium Groupon, and hard Horse. This is reflected in Table\ref{tab:correlation} where this order is the exact same as ATE deviation. Although there may be more work to do on combining SMD and A2A, this metric is ready to use and can have a significant impact on PSM evaluation.

%\textbf{Best PSM methods.} While this is not the primary objective of our study, we conducted comparisons of matching methods on our tasks. Our main finding is that applying logit on predicted probabilities results in models that are considered invalid based on the SMD. This also applies to \emph{nearest} matching in MatchIt. Ultimately, for Python practitioners, PsmPy with CLR stands out as the best method, while ElasticNet optimal is the top choice for R users.

\paragraph{Conclusion.}Propensity-based methods can compensate for selection bias based on given covariates. We have shown that SMD effectively evaluates these methods for this specific purpose. Incidentally, they also provide good compensation for causal effect estimation when the covariates involved in outcome estimation are the same as those in selection, \ie, they are confounders. However, when covariates are not confounders or in the presence of mixed effects, SMD may fail to select the best method, and propensity correction might be inadequate.

This study introduces A2A, a new metric that offers insight into bias correction by evaluating the entire matching pipeline, including target effect estimation, on the data at hand. A2A evaluates a correction method based on its outcome correction capability, which, to our knowledge, is a novel approach. In scenarios with a low number of confounders, A2A is superior to SMD for selecting the best matching method. Since the number of confounders can be hypothesized but not definitively determined, it is challenging to identify the most accurate metric. To address this, we propose combining them with a Pareto optimal selection to leverage the strengths of both metrics and minimize estimation error.

\paragraph{Limitations.} Although promising, our study would benefit from replication on a broader range of datasets to generate more insights and improve the existing pipeline, in particular by replacing DBScan by a simpler and more interpretable method.
Lastly, this study lays the groundwork for future research. If this analysis is limited to PSM, the same strategy can be applied on other bias correction techniques such as Inverse Propensity Weighting. Additionally, A2A currently functions as a relative metric for refining PSM selection, necessitating the computation of all pipelines to identify the best ones. Our goal is to develop an absolute metric usable with a threshold, similar to SMD. Furthermore, the scheme behind A2A could potentially be adapted to find weights, complementary to propensity, that allows for a better correction of the ATE.

\begin{credits}
\subsubsection{\ackname}
This research was conducted on our spare time. Alexandre Abraham owes his deepest gratitude to his family, Hinata and Audrey for providing him with the time to pursue this work. He also thanks Implicity for indirectly contributing through the experiment that inspired this work.

\subsubsection{\discintname}
The authors have no competing interests to declare that are relevant to the content of this article.
\end{credits}
%
% ---- Bibliography ----
%
% BibTeX users should specify bibliography style 'splncs04'.
% References will then be sorted and formatted in the correct style.
%
\bibliographystyle{splncs04}
\bibliography{paper}
%% Note that this preceding line implies that you store your BibTeX references in a file called 'mybibliography.bib'. If you instead store your references in a file with a different name, for instance 'references.bib', the preceding line should read '\bibliography{references}'. Whatever you do, DO NOT put the file name extension .bib inside the \bibliography command; this will trip up LaTeX compilers. 
%

\newpage

\appendix

\renewcommand{\thealgocf}{\thesection.\arabic{algocf}}
\renewcommand\thefigure{\thesection.\arabic{figure}}

\section{Algorithms}
Let $\mathbf{x}_j$ be the $j$th row of $X$.
\begin{algorithm}
  %\caption{Hill-climbing clustering}
  \caption{Selecting subsets to match by Hill-Climbing clustering}
  \label{alg:clustering}
  
  \KwData{Data $(\left\{X_0, X_1\right\},\, Y)$, cluster membership probabilities $p \in [0, 1]^2$, loss function $\mathcal{L}$, number of iterations $K$.}
    
  $\mathbf{c}_j$ \textleftarrow Assign $\mathbf{x}_j \in X_0$ to a cluster at random following $p$, $\forall j \in \{0, \ldots, |X_0|\}$\;

  $Z^{h} \gets \left\{\mathbf{x}_j \in X_0 :\ \mathbf{c}_j=h,\, \forall j \in \{0, \ldots, |X_0|\}  \right\}$, where $h=\{0, 1\}$\;
  
  $l^{\textsf{current}}$ \textleftarrow $\mathcal{L}\left(X,\, Y,\, \left\{Z^{(0)}, Z^{(1)} \right\}\right)$ \quad (evaluate using Eq.~\ref{eqn:main_loss})\;
  \For{$i\leftarrow 1$ \KwTo $K$}{
  
    %$j, k$ \textleftarrow Random indices in $X_0$ with $\mathbf{c}_j \neq \mathbf{c}_k$
    
    $j, k$ \textleftarrow Draw $(j, k) \in \{0, \ldots |X_0|\}^2$ at random with $\mathbf{c}_j \neq \mathbf{c}_k$    

    %\textsf{clusters}[j], \textsf{clusters}[k] \textleftarrow \textsf{clusters}[k], \textsf{clusters}[j]
    $\mathbf{c}_j, \mathbf{c}_k \gets \mathbf{c}_k, \mathbf{c}_j$ \quad (swap cluster assignment)

  $Z^{h} \gets \left\{\mathbf{x}_j \in X_0 :\ \mathbf{c}_j=h,\, \forall j \in \{0, \ldots, |X_0|\}  \right\}$, where $h=\{0, 1\}$\;

    $l^\textsf{new} \gets \mathcal{L}\left(X,\, Y,\, \left\{Z^{(0)}, Z^{(1)}\right\}\right)$
    
    \eIf{$l^\textsf{new} < l^\textsf{current}$}{
        $l^\textsf{current} \gets l^\textsf{new}$
    }
    {
        $\mathbf{c}_j, \mathbf{c}_k \gets \mathbf{c}_k, \mathbf{c}_j$ \quad (swap cluster assignment)
    }
}
    %$\left\{X_0^{(0)}, X_0^{(1)}\right\} \gets \left\{Z^{(0)}, Z^{(1)}\right\}$\;
  \Return{$\left\{Z_0^{(0)}, Z_0^{(1)}\right\}$ \quad (selected subsets/clusters)}\;
\end{algorithm}

\vfill
\newpage
\section{SMD vs. A2A}

This figure illustrates the trade-off between the two metrics. It's challenging to determine the superior method among Bipartify LR, and GLM/GAM solely from this plot. Hence, we opt for the Pareto optimal selection of methods. PsmPy's RF could potentially be considered Pareto-optimal, but it is effectively eliminated because its SMD exceeds 10%.

\begin{figure}
\includegraphics[width=\textwidth]{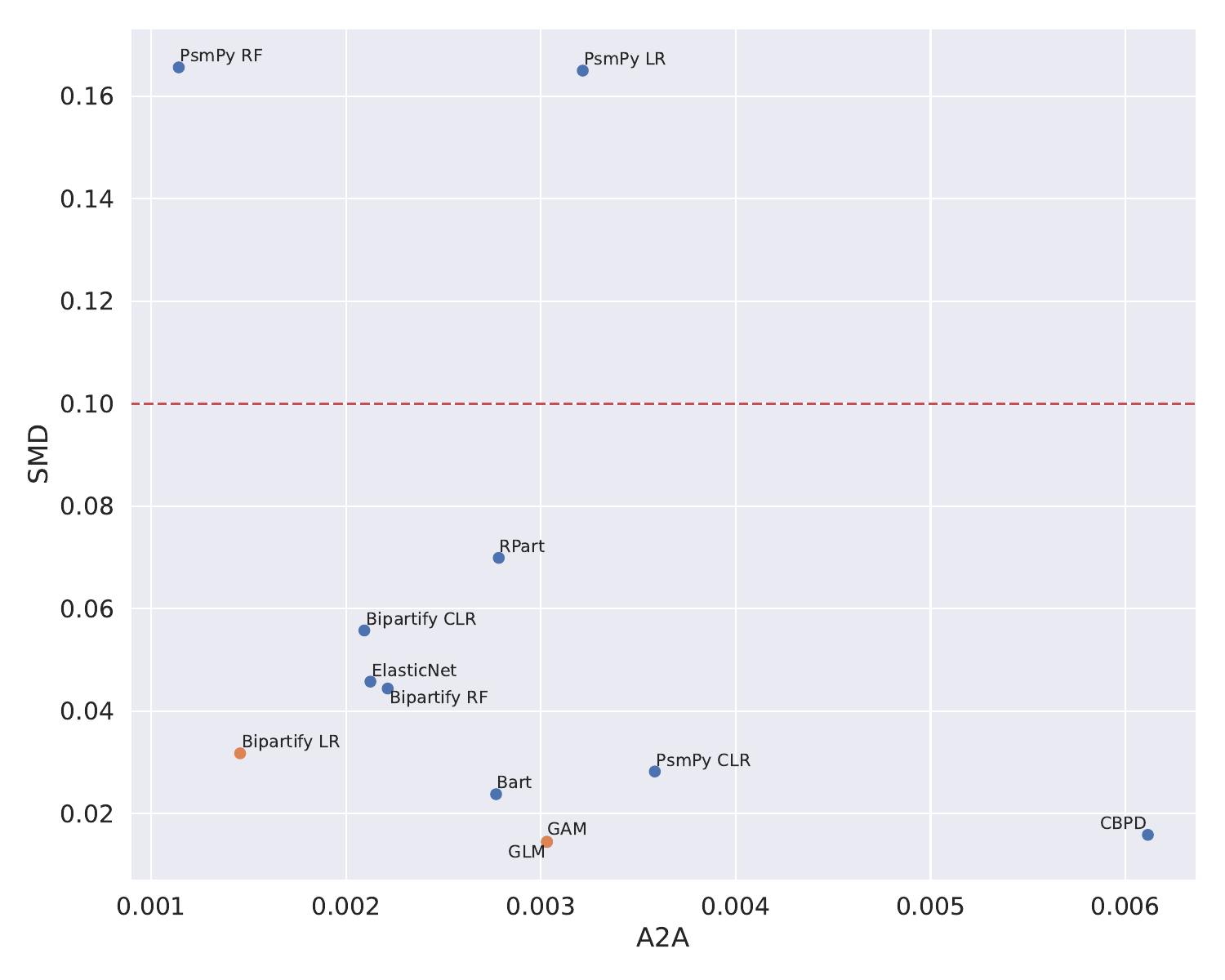}

\caption{Values for SMD and A2A for the task NHANES. Methods above 0.1 in SMD are invalid. Bipartify LR and GLM form the Pareto optimum according to both metrics.}
\label{fig:smda2a}

\end{figure}

\end{document}